\documentclass{nesy2026} 

\jmlrproceedings{}{}
\jmlrvolume{}
\jmlryear{}
\jmlrpages{}

\usepackage{amsmath}
\usepackage{booktabs}

\newcommand{\std}[1]{{\scriptsize $\pm$#1}}
\newcommand{\hquot}{/\!\!/}        
\newcommand{\U}{\mathcal{U}}        
\newcommand{\Id}{\mathrm{Id}}
\newcommand{\Aut}{\mathrm{Aut}}

\newcommand{\Stab}{\mathrm{Stab}}
\newcommand{\Sym}{\mathrm{Sym}}

\title[Homotopy-theoretic neurosymbolic inference]{A homotopy-type-theoretic
generalization of neurosymbolic inference}

\clearauthor{\Name{Fernando Zhapa-Camacho}
  \Email{fernando.zhapacamacho@kaust.edu.sa}\\
  \Name{Robert Hoehndorf} \Email{robert.hoehndorf@kaust.edu.sa}\\
\addr Computer Science Program, Computer, Electrical and Mathematical Sciences and Engineering Division,
King Abdullah University of Science and Technology (KAUST), Thuwal, Saudi Arabia}

\begin{document}

\maketitle

\begin{abstract}
  A wide range of neurosymbolic
  (NeSy) systems compute one functional: a belief-weighted sum of a
  logical quantity over a space of $\sigma$-structures, of which
  weighted model counting, fuzzy logic, and probabilistic logic are
  special cases. This account is built on \emph{sets}, and a set
  deliberately forgets two things that are important for NeSy: when
  two $\sigma$-structures are the same up to a symmetry of the theory,
  and how many distinct proofs witness a query. \emph{Types}, in the
  sense of homotopy type theory, preserve this information and turn
  the functional into a belief-weighted homotopy cardinality, a notion
  of size that counts each object in inverse proportion to its
  symmetries. We develop the framework from scratch for NeSy systems,
  prove a conservativity theorem that recovers the classical
  functional when symmetries are trivial, and show that the symmetry
  our framework exposes is exactly the one behind reasoning
  shortcuts. The payoff is concrete: the
  shortcut-aware concept posterior that recent methods reach by
  ensembling or expressive density estimation is the only
  symmetry-invariant point of the confusion-set simplex, computable in
  closed form by averaging a single model over the symmetry group. On
  MNIST reasoning-shortcut benchmarks this single-model wrapper is
  better calibrated than a diversity-trained ensemble, while leaving
  label accuracy and identifiable concepts untouched. Code is freely
  available at \url{https://github.com/bio-ontology-research-group/hott-nesy}.
\end{abstract}

\section{Introduction}
\label{sec:intro}

Much of neurosymbolic AI can be read as a single recipe: a neural
network proposes weights over possibilities, a logical theory says
which possibilities are admissible, and inference aggregates the
weights of the admissible ones \citep{desmet2025defining}. Write
$\Omega$ for a space of \emph{$\sigma$-structures} (assignments of
values to the symbols of a logical language), $\ell$ for a \emph{logic
  function} giving the value of a sentence $\varphi$ in a
$\sigma$-structure, and $b_\theta$ for a neural \emph{belief} over
$\sigma$-structures. Neurosymbolic inference is then the
belief-weighted aggregate
\begin{equation}
\label{eq:deraedt}
F_\theta(\varphi) \;=\; \int_{\Omega} \ell(\varphi,\omega)\, b_\theta(\omega)\,\mathrm{d}m(\omega).
\end{equation}
Here $m$ is the base measure on $\Omega$. With the counting measure,
Eqn.~\eqref{eq:deraedt} reduces to weighted model counting (WMC); over a
continuous domain it becomes weighted model integration. Through the lens
of algebraic model counting \citep{kimmig2017algebraic}, varying the
arithmetic used to combine values recovers probabilistic, fuzzy,
tropical (MaxSAT/Viterbi), and provenance inference; varying the
belief recovers the semantic loss \citep{xu2018semantic}, DeepProbLog
\citep{manhaeve2018deepproblog}, logic
tensor networks \citep{badreddine2022logic}, Scallop
\citep{huang2021scallop}, and more. Equation \eqref{eq:deraedt} is
therefore a common denominator for large parts of the NeSy field.

A space of $\sigma$-structures is a \emph{set}, and a set is a
deliberately forgetful object: it records only which elements it contains
but not \emph{(i)} that two
$\sigma$-structures are ``the same up to a symmetry'' of the theory,
nor \emph{(ii)} how many distinct derivations witness that a query
holds. Both are central to relational and structured NeSy. If two
individuals are interchangeable (no evidence distinguishes them), the
$\sigma$-structures obtained by swapping them are not different ones,
yet Eqn.~\eqref{eq:deraedt} counts them as two; if a query is provable
in several ways, Eqn.~\eqref{eq:deraedt} collapses this to a single
number. The first
symmetry is not
an exotic concern: it is exactly the structure behind \emph{reasoning
  shortcuts} \citep{marconato2023notall}, a central failure mode of
NeSy learning.

We investigate what Eqn.~\eqref{eq:deraedt} becomes if its underlying sets
are replaced by \emph{types}, in the sense of homotopy type theory
(HoTT) \citep{hottbook}. A type is like a set that additionally
remembers, for any two elements, the ways in which they may be
identified; from this it recovers each element's symmetry group and
each proof's identity. The correct
notion of \emph{size} for such an object is not cardinality but
\emph{homotopy cardinality}
\citep{baez2001finite,leinster2008euler},
which counts each element weighted by the reciprocal of its
symmetries. Carried through this change of
foundations (Section~\ref{sec:lift}), Eqn.~\eqref{eq:deraedt} becomes a
belief-weighted homotopy cardinality that retains this information and
specializes back to the original when all symmetries are trivial.

We keep the development self-contained and assume no background in
type theory: Section~\ref{sec:hott} introduces every notion we use,
and Appendix~\ref{app:hott} expands it in logic and knowledge-graph
terms. Our contributions are:
\begin{itemize}
\item the lift of each ingredient of Eqn.~\eqref{eq:deraedt} one
  level, exposing two parameters the classical functional cannot name:
  a \emph{symmetry parameter} and a \emph{truncation level};
\item a \emph{conservativity theorem} (Section~\ref{sec:theorem}),
  verified in Lean~4 with Mathlib~\citep{mathlib}: the belief-weighted
  homotopy cardinality is a symmetry-corrected weighted count, and
  Eqn.~\eqref{eq:deraedt} returns when the symmetry is trivial or the
  semiring idempotent;
\item the identification (Section~\ref{sec:rs}) of the symmetry parameter
  with the mechanism behind reasoning shortcuts, and the closed-form,
  single-model method for shortcut-aware concept posteriors it yields
  (\emph{orbit-averaging});
\item experiments (Section~\ref{sec:exp}) on MNIST reasoning-shortcut
  tasks where the wrapper is better calibrated than ensemble and
  training-time baselines, with accuracy and identifiable concepts
  preserved;
\item a generating-function consequence for exchangeable models
  (Section~\ref{sec:examples}), what types buy (Section~\ref{sec:impact}),
  and the route to the continuous side through cohesion
  (Section~\ref{sec:cohesion}).
\end{itemize}

\section{Background}
\label{sec:background}

\subsection{The neurosymbolic inference functional}

As shown in Eqn.~\eqref{eq:deraedt}, a NeSy model
\citep{desmet2025defining} fixes a signature $\sigma$ of symbols with
associated domains, so that a \emph{$\sigma$-structure}
$\omega\colon\sigma\to D$ assigns each symbol a value; the space of
all $\sigma$-structures is $\Omega=D^{\sigma}$. A semantics sends a
sentence $\varphi$ and a $\sigma$-structure $\omega$ to a truth value
in a set $V\hookrightarrow\mathbb{R}^{+}$ (Boolean $\{0,1\}$, fuzzy
$[0,1]$, etc.), and the logic function $\ell(\varphi,\omega)$
reads off the part we wish to aggregate. A belief
$b_\theta\colon\Omega\to\mathbb{R}^{+}$, typically computed by a neural
network (the \emph{perception}) from the raw input, weights
$\sigma$-structures. For finite $\Omega$ with the counting
measure, Eqn.~\eqref{eq:deraedt} is the weighted model count
\begin{equation}
\label{eq:wmc}
F_\theta(\varphi)\;=\;\sum_{\omega\models\varphi} b_\theta(\omega).
\end{equation}

\begin{example}[A two-individual smokers program]
\label{ex:running}
Let two individuals
$1,2$ each either smoke or not, so a $\sigma$-structure is a function
$\omega\colon\{1,2\}\to\{\bot,\top\}$ and $\Omega$ has four
elements. Suppose perception outputs an independent smoking
probability $p_i$ for each individual, giving the belief
$b_\theta(\omega)=\prod_i
p_i^{[\omega(i)=\top]}(1-p_i)^{[\omega(i)=\bot]}$. For the query
$\varphi=$ ``someone smokes'', Eqn.~\eqref{eq:wmc} sums $b_\theta$ over the
three satisfying $\sigma$-structures and gives $1-(1-p_1)(1-p_2)$. We
return to this program throughout.
\end{example}

A single algebraic generalization underlies the field's diversity:
replacing the value set $V$ by a commutative semiring
$(\oplus,\otimes)$ and reading Eqn.~\eqref{eq:wmc} in that semiring is
\emph{algebraic model counting}
\citep{kimmig2017algebraic}: probability uses $(+,\times)$, fuzzy
logic $(\max,\min)$, MaxSAT and Viterbi the tropical
$(\max,+)$, provenance free semirings of proof terms.

These readings are all \emph{decategorifications}: passages from a
structured object to a numerical invariant that forgets the
structure. The prototype is that the natural numbers are
the decategorification of finite sets, where a number records an
isomorphism class, disjoint union becomes $+$, and Cartesian product
becomes $\times$; algebraic model counting then varies which invariant
one reads off (Section~\ref{sec:lift}).

\subsection{Reasoning shortcuts}
\label{sec:rs-bg}

A NeSy model supervises its perception, which predicts the latent
\emph{concepts} of each input, only through the logical output, and
this under-determination has a well-studied failure mode. Write
$\beta$ for the map from a concept assignment to the program output. A
\emph{reasoning shortcut} is a perception that minimizes the training
loss while assigning its concepts the wrong meaning, induced by a
relabeling $\alpha$ that leaves the output unchanged,
$\beta(\alpha(c))=\beta(c)$
\citep{marconato2023notall,bortolotti2025identifiability}. The concept
assignments reachable by such relabelings form a \emph{confusion set}
\citep{vankrieken2025rsindependence}, and, under mild assumptions, the
loss-minimizing perceptions are convex combinations of the
deterministic optima \citep{marconato2023notall}. Under the common
\emph{independence assumption} (the perception factorizes over
concepts, \citealp{vankrieken2024independence}), the model commits to
a single element of each confusion set with no evidence to prefer it.

The concepts within a confusion set are \emph{unidentifiable} from the
data, so a predictor should be \emph{uncertain} across them: the
desired \emph{shortcut-aware} posterior is the uniform,
maximum-entropy distribution on each confusion set
\citep{marconato2024bears}. Existing methods reach it with a
diversity-trained ensemble (\textsc{Bears},
\citealp{marconato2024bears}), a uniform-target regularizer
\citep{vankrieken2025rsindependence}, or an expressive joint
distribution over concepts \citep{vankrieken2025nesydm}.
Section~\ref{sec:rs} recovers this target in closed form by reading
confusion sets as symmetry orbits.

\subsection{Types, identifications, and symmetry}
\label{sec:hott}
\label{sec:types}

A type is a set that also remembers how its elements can be
identified. For neurosymbolic inference the elements are
$\sigma$-structures, and an identification is a renaming of
individuals carrying one $\sigma$-structure to another that respects
the theory: an isomorphism of $\sigma$-structures, or an
automorphism when the two coincide.

For a type $A$ and elements $a,b:A$ there is an \emph{identity type}
$\Id_A(a,b)$ whose elements are \emph{identifications} of $a$ with
$b$. Take $A$ to be the $\sigma$-structures of a theory $T$: an
identification of $a$ with $b$ is a renaming of the domain
individuals that carries $a$ to $b$ and respects $T$. There may be
none (the two are genuinely different models), exactly one (the
situation in an ordinary set of $\sigma$-structures), or several. The
self-identifications of one $\sigma$-structure compose and invert, so
they form a group: the \emph{automorphism group}
$\Aut(a)$. This is the information a set discards, being the
degenerate case in which the only
renaming fixing any $\sigma$-structure is the identity.

In the smokers program of Example~\ref{ex:running}, if no evidence
distinguishes individuals $1$ and $2$, then swapping them identifies
the $\sigma$-structure ``only $1$ smokes'' with ``only $2$ smokes''.
The same swap leaves ``nobody smokes'' and ``everybody smokes''
unchanged, so it is a nontrivial automorphism of each. The type of
$\sigma$-structures records this symmetry; the set cannot.

\subsection{Truncation: propositions, sets, groupoids}
\label{sec:truncation}

Types are stratified by how much identification structure they carry,
the \emph{truncation level}. A \emph{$(-1)$-type}, or
\emph{proposition}, has at most one element, a yes/no fact such as
whether a ground atom holds. A \emph{$0$-type}, or \emph{set}, has at
most one identification between any two elements, and the
$\sigma$-structures $\Omega=D^{\sigma}$ form one. A \emph{$1$-type}, or
\emph{groupoid}, may have nontrivial automorphism groups but no higher
structure, and $\sigma$-structures carrying their renaming symmetries
live here.

\subsection{Dependent sums and homotopy cardinality}
\label{sec:sigma}

If to each $a:A$ we assign a type $P(a)$, the \emph{dependent sum}
$\sum_{a:A}P(a)$ is the type of pairs $(a,p)$ with $p:P(a)$. We take
$P(\omega)$ to be the \emph{derivations} that witness the query in
$\sigma$-structure $\omega$: a Boolean once we forget which, or the
several distinct proof trees that provenance records. Its
identifications combine renamings of
models with identifications of the derivations they carry, so
symmetries of models and of proofs are tracked together.

Ordinary counting is the wrong measure of \emph{size} for a groupoid
of $\sigma$-structures, because it ignores symmetry: a
$\sigma$-structure fixed by a nontrivial renaming should count as less
than a whole one. The right notion is \emph{homotopy cardinality}
\citep{baez2001finite,leinster2008euler}, which counts each element in
inverse proportion to its symmetries,
\begin{equation}
\label{eq:hcard}
\lvert X\rvert \;=\; \sum_{[x]} \frac{1}{\lvert\Aut(x)\rvert},
\end{equation}
the sum running over $\sigma$-structures up to renaming. A set of $k$
$\sigma$-structures has $\lvert X\rvert=k$;
a single model with symmetry group $G$ counts as $\tfrac{1}{\lvert G\rvert}$.
Homotopy cardinality requires finiteness (finitely many isomorphism
classes, finite automorphism groups; ``$\pi$-finiteness'').

To connect this to a group of renamings, we use three standard notions
from group actions. When a finite group $G$ acts on a finite set $X$
of $\sigma$-structures, the \emph{orbit}
$G\cdot x=\{g\cdot x:g\in G\}$ collects the structures
indistinguishable from $x$ up to renaming,
the \emph{stabilizer} $\Stab_G(x)=\{g\in G:g\cdot x=x\}$ collects the
renamings fixing $x$, and each orbit has size
$\tfrac{\lvert G\rvert}{\lvert\Stab_G(x)\rvert}$ (the
\emph{orbit--stabilizer theorem}). The \emph{action
  groupoid} $X\hquot G$ adds an identification $x\to g\cdot x$ for
every $g\in G$; the double slash distinguishes it from the orbit set
$X/G$, which forgets the stabilizers. The two preceding facts give its
homotopy cardinality,
\begin{equation}
\label{eq:actcard}
\lvert X\hquot
G\rvert=\sum_{\text{orbits}}\frac{1}{\lvert\Stab_G(x)\rvert}=\frac{\lvert
  X\rvert}{\lvert G\rvert}. 
\end{equation}

\begin{example}[Models up to relabeling]
\label{ex:hcard}
Take the four $\sigma$-structures of Example~\ref{ex:running} with
renaming group $\Sym(2)=\{\mathrm{id},\text{swap}\}$. ``Nobody'' and
``everybody'' are each fixed by the swap (stabilizer of size $2$); the
two ``exactly one'' models form one orbit with trivial stabilizer.
Then Eqn.~\eqref{eq:hcard} gives $\tfrac12+\tfrac12+1=2=\tfrac{4}{2}$,
the symmetry-weighted count, not the number $3$ of
models-up-to-relabeling.
\end{example}

\section{Lifting the functional}
\label{sec:lift}

We now carry each ingredient of Eqn.~\eqref{eq:deraedt} from sets to
types (Table~\ref{tab:lift}); in every row the classical object is
recovered by truncating.

\begin{table}[t]
\centering\small
\begin{tabular}{@{}lll@{}}
\toprule
\citet{desmet2025defining} & our work & recovered when\\
\midrule
structure set $\Omega$ & structure type $\Omega$ ($\Id_\Omega$ =
                         symmetries) & $\Omega$ a set\\
value $\ell(\varphi,\omega)\in V$ & proof family $P\colon\Omega\to\U$
                                        & $P(\omega)$ a proposition\\
integral $\int \ell\, b\,\mathrm{d}m$ & homotopy card.\ of
                                        $\sum_\omega P(\omega)$ &
                                                                  trivial symmetry\\
semiring $V$ & valuation $\lVert\cdot\rVert_V$ & choice of semiring\\
\bottomrule
\end{tabular}
\caption{Lifting the four ingredients of neurosymbolic inference one level.}
\label{tab:lift}
\end{table}

First, the structures become a type $\Omega:\U$ whose
identity types record structure symmetries, so that $\sigma$-structures
related by an isomorphism are identified; the classical
$\Omega=D^{\sigma}$ is the $0$-truncated case. Second, the truth value
becomes a proof family: we replace $\ell(\varphi,\omega)$ by a type
family $P\colon\Omega\to\U$, where $P(\omega)$ is the type of
\emph{witnesses} that $\varphi$ holds at $\omega$. Its truncation level
is a parameter, so that a proposition records only \emph{whether}
$\varphi$ holds, a set records \emph{how many} derivations witness it
(provenance), and a groupoid
identifies derivations related by a symmetry. Third, inference becomes
a homotopy cardinality: the dependent sum
$\sum_{\omega:\Omega}P(\omega)$ is the type of \emph{(structure,
  witness)} pairs, and its belief-weighted homotopy cardinality
\begin{equation}
\label{eq:functional}
N(P,b)\;=\;\Big\lVert \textstyle\sum_{\omega:\Omega}P(\omega)\Big\rVert_{V,b}
\end{equation}
is our inference value. When $\Omega$ is a set, $P$ a proposition, and
$\lVert\cdot\rVert_{V,b}$ the belief-weighted counting measure,
Eqn.~\eqref{eq:functional} collapses to the weighted model count
\eqref{eq:wmc}. Fourth, the valuation $\lVert\cdot\rVert_V$ turning
the proof type into a number is
exactly the choice of commutative semiring: algebraic model
counting \citep{kimmig2017algebraic} viewed as decategorification maps
applied to the \emph{same} object \eqref{eq:functional}.

The lift exposes two parameters Eqn.~\eqref{eq:deraedt} cannot
express. The first is a \emph{symmetry parameter}: one may quotient
$\Omega$ by any subgroup $G\le\Aut(T)$ of the theory's symmetries,
interpolating between $G=\mathbf{1}$ (every grounding distinct) and
$G=\Aut(T)$ ($\sigma$-structures counted up to isomorphism). This
parameter is not free, because Theorem~\ref{thm:main} needs the belief to
be $G$-invariant, so $G$ is exactly the symmetry shared by the theory
and the belief: a perception that scores individuals differently
breaks the symmetry and forces $G=\mathbf{1}$, whereas an exchangeable
perception permits the full
$G=\Aut(T)$. The second is a \emph{truncation level}: the level of $P$
separates systems that check satisfaction (propositions; e.g.\
\citealp{badreddine2022logic}) from those that count derivations
(sets; e.g.\ \citealp{manhaeve2018deepproblog,huang2021scallop}) from
those that identify symmetric derivations (groupoids).

On the smokers program (Example~\ref{ex:running}) with query ``someone
smokes'', the structure--witness object $\sum_{\omega}P(\omega)$ holds
the three satisfying models with their beliefs. The probability
semiring $(+,\times)$ sums these to $1-(1-p_1)(1-p_2)$, while the
tropical $(\max,+)$ reading of the \emph{same} object returns the most
probable satisfying model. The object is fixed; the semiring
chooses which number to read off it.

\subsection{The conservativity theorem}
\label{sec:theorem}

We make precise, and machine-check, that Eqn.~\eqref{eq:functional}
extends Eqn.~\eqref{eq:deraedt} faithfully, working in the finite,
$1$-truncated (groupoid) case, where the structure--witness type with
symmetry is a finite action groupoid (Section~\ref{sec:sigma}).

\begin{definition}[Belief-weighted groupoid cardinality]
\label{def:gcard}
Let a finite group $G$ act on a finite type $X$ of structure--witness
pairs, and let $b\colon X\to\mathbb{Q}^{+}$ be \emph{$G$-invariant},
$b(g\cdot x)=b(x)$. The belief-weighted homotopy cardinality of
$X\hquot G$ is
$\lvert X\hquot G\rvert_{b}=\sum_{[x]\in X/G}
\tfrac{b(x)}{\lvert\Stab_G(x)\rvert}$.
\end{definition}

\begin{theorem}[Conservativity bridge]
\label{thm:main}
With the notation of Definition~\ref{def:gcard},
$\lvert X\hquot G\rvert_{b}=\tfrac{1}{\lvert G\rvert}\sum_{x\in X}
b(x)$.
\end{theorem}

The proof is orbit--stabilizer arithmetic (Appendix~\ref{app:proofs}),
formalized in Lean~4 over Mathlib (Appendix~\ref{app:lean}). When $G$
is trivial and $P$ is a proposition,
$N(P,b)=F_\theta(\varphi)$ is the
weighted model count of Eqn.~\eqref{eq:wmc}. Theorem~\ref{thm:main} is the
$(+,\times)$ instance; an arbitrary commutative semiring admits a
second, dual degeneracy.

\begin{proposition}[Two degeneracies of the lift]
\label{prop:inert}
Read Eqn.~\eqref{eq:functional} in a commutative semiring
$(\oplus,\otimes)$ via $\lVert\cdot\rVert_{V}$, with $b$ $G$-invariant.
The lifted value equals the set-based one,
$\lvert X\hquot G\rvert_{b}=\bigoplus_{x\in X} b(x)$, under either
degeneracy: \emph{(a)} the symmetry is trivial, $G=\mathbf{1}$ (any
semiring); or \emph{(b)} $\oplus$ is idempotent (any $G$).
\end{proposition}

Case (a) is Theorem~\ref{thm:main} at $G=\mathbf{1}$. In case (b) the
belief is constant on each orbit, so an idempotent $\oplus$ fuses the
orbit's equal copies into one ($v\oplus\cdots\oplus v=v$) while the
$\tfrac{1}{\lvert\Stab\rvert}$ weight is the identity
(Appendices~\ref{app:proofs} and~\ref{app:lean}). The degeneracies are dual: trivial symmetry
has nothing to quotient, an idempotent semiring nothing to overcount.
The lift is therefore visible only in the non-idempotent semirings
(probabilistic $(+,\times)$, the gradient semiring, counting,
provenance) and inert on the idempotent ones, such as $(\max,+)$
(MPE/Viterbi), $(\max,\min)$ (fuzzy), and $(\vee,\wedge)$ (SAT); the
most-probable-$\sigma$-structure reading is symmetry-invariant for
free. Remark~\ref{rem:divisible} discusses which semirings the
$\tfrac{1}{\lvert\Aut\rvert}$ weight is defined in.

\subsection{Reasoning shortcuts as residual symmetry}
\label{sec:rs}

The symmetry parameter is the structure behind
reasoning shortcuts (Section~\ref{sec:rs-bg}), and it yields a
practical method.  The output-preserving relabelings are invertible and closed under
composition: a group, the symmetry parameter $G\le\Aut(T)$
instantiated by ``preserves the belief''. A confusion set is then exactly an orbit
$G\cdot x$ under this group, and two facts the
literature states set-theoretically become orbit--stabilizer
arithmetic: the number of distinct shortcuts through a
$\sigma$-structure is the orbit size
$\tfrac{\lvert G\rvert}{\lvert\Stab\rvert}$, and the symmetry-corrected
functional $\lvert X\hquot G\rvert_{b}$ is invariant under this group
by construction (Theorem~\ref{thm:main}).  The independence assumption
is the trivial case $G=\mathbf 1$: the loss minima are the
disconnected deterministic choices of Section~\ref{sec:rs-bg}
\citep{vankrieken2024independence}.

The shortcut-aware posterior is the only symmetry-invariant point.
Concept posteriors on a confusion set, an orbit, form a probability
simplex whose vertices the group action permutes, so the single
invariant point is the uniform distribution: the maximum-entropy
target of Section~\ref{sec:rs-bg}. Here it is forced. For any
belief the symmetry parameter fixes, the orbit-conditional posterior
is uniform, $\tfrac{b(y)}{\sum_{z}b(z)}=\tfrac{1}{\lvert G\cdot x\rvert}$, the same
invariant measure the functional integrates against (proof in Lean,
Appendix~\ref{app:lean}).

This response lives only in a non-idempotent semiring. Read in the $(\max,+)$ of
most-probable-model inference, a confusion set's tied values
collapse to a single number (Proposition~\ref{prop:inert}), and the
aware uniform posterior becomes indistinguishable from an overconfident
commitment to one shortcut.
Orbit-averaging reaches this invariant point in closed form, with
neither an ensemble nor a density model: \emph{symmetrize} any base
perception $p_\theta(c\mid x)$ over the output-preserving group $G$,
\begin{equation}
\label{eq:reynolds}
\bar p(c\mid x)\;=\;\frac{1}{\lvert G\rvert}\sum_{g\in G} p_\theta(g^{-1}\!\cdot c\mid x).
\end{equation}

\begin{proposition}[Orbit-averaging is closed-form shortcut awareness]
  \label{prop:method}
  Let $G$ be output-preserving. The symmetrized posterior of
  Eqn.~\eqref{eq:reynolds} then has three properties. \emph{(i)} It
  leaves the induced label distribution unchanged, so label accuracy
  is preserved exactly. \emph{(ii)} It equals the uniform
  (maximum-entropy) distribution on each confusion set, the only
  $G$-invariant point. \emph{(iii)} It is computed in one forward pass
  plus a sum over $G$, with the orbit size
  $\tfrac{\lvert G\rvert}{\lvert\Stab\rvert}$ as a per-input
  shortcut-degeneracy score.
\end{proposition}

\noindent Part (i) holds because each $g$ permutes every label
fiber $\beta^{-1}(y)$ within itself
(Appendix~\ref{app:proofs}); part (ii) is the invariant-point
statement above, and part (iii) is Eqn.~\eqref{eq:reynolds} read
literally. So orbit-averaging keeps confident, correct
predictions on identifiable concepts (orbit size $1$) and is
calibrated-uncertain exactly on the confusable ones, from a single
model. The group $G$ is read off the knowledge, and computing it is
graph-automorphism detection (encode the theory as a vertex-colored
graph), practical at scale with nauty \citep{mckay2014practical} or
saucy \citep{darga2008faster}. The closed form is exact when $G$ is a genuine
symmetry of the data;
where it is only partial, an ensemble \citep{marconato2024bears} or
expressive model
\citep{vankrieken2025nesydm} represents the rest of the simplex
(Appendix~\ref{app:limitations}). To
our knowledge, the reading of
reasoning shortcuts as orbits of the theory's symmetry group, and the
closed-form awareness method it yields, are new.

\section{Experiments}
\label{sec:exp}

We test orbit-averaging on MNIST reasoning-shortcut tasks, in the
style of the \textsc{rsbench} suite \citep{bortolotti2024rsbench},
where the
symmetry group is known and concept ground truth is available.
Appendix~\ref{app:exp} gives full details, architectures, and a second
task (XOR-parity, group $\mathbb{Z}/2$).

\paragraph{Setup.}
A single MNIST digit in $\{0,\dots,5\}$ is mapped by a knowledge
$\beta$ that merges labels ($0\!\mapsto\!0$, $\{1,2\}\!\mapsto\!1$,
$\{3,4,5\}\!\mapsto\!2$); training sees only $\beta$. The
output-preserving group is
$G=\Sym\{0\}\times\Sym\{1,2\}\times\Sym\{3,4,5\}$, so digit $0$ is
identifiable (orbit size $1$) while $\{1,2\}$ and $\{3,4,5\}$ are
confusable (orbit sizes $2,3$): a reasoning shortcut on the confusable
digits only. We compare a base model, the same model orbit-averaged
\eqref{eq:reynolds}, the RS-independence loss
\citep{vankrieken2025rsindependence}, \textsc{NeSyDM}
\citep{vankrieken2025nesydm}, and a
diversity-trained \textsc{Bears}
ensemble \citep{marconato2024bears}, whose diversity penalty
Appendix~\ref{app:exp} states in full.

Orbit-averaging receives the true group $G$, read off the knowledge
$\beta$ that the task fixes by construction (Appendix~\ref{app:exp}),
and the RS-independence loss supervises toward the uniform
distribution on the fibers of the same $\beta$, on this task the
orbits of $G$. \textsc{Bears} and \textsc{NeSyDM} use $\beta$ only
through the label likelihood and reach awareness by training pressure
alone. The comparisons therefore measure what explicitly using the
symmetry buys, not whether it can be discovered from data
(Appendix~\ref{app:limitations}). Appendix~\ref{app:ablation} ablates
$G$ itself: a misspecified group shows up in validation label
accuracy, not in calibration error.

We
report four quantities, defined fully in Appendix~\ref{app:exp}: the
accuracy of the merged label $\beta$ (\emph{label}), the tie-aware
expected calibration error (ECE, the gap between confidence and
accuracy) of the digit posterior on the identifiable digit
(\emph{id-ECE}) and on the confusable digits (\emph{conf-ECE}), and
the mean posterior entropy on the confusable digits
(\emph{conf-ent.}), which a shortcut-aware model keeps high.

\begin{table}[t]
\centering\small
\begin{tabular}{@{}lcccccc@{}}
\toprule
method & models & epochs & label & id-ECE & conf-ECE & conf-ent.\\
\midrule
base & 1 & 4 & $0.996$\std{0.001} & $0.002$\std{0.001} & $0.588$\std{0.109} & $0.05$\std{0.04}\\
\textbf{ours (orbit-avg.)} & \textbf{1} & \textbf{4} & $0.996$\std{0.001} & $0.004$\std{0.002} & $\mathbf{0.013}$\std{0.008} & $0.94$\std{0.00}\\
RS-independence & 1 & 4 & $0.995$\std{0.001} & $0.004$\std{0.002} & $0.054$\std{0.034} & $0.94$\std{0.00}\\
\textsc{NeSyDM} & 1 & 4 & $0.993$\std{0.002} & $0.008$\std{0.006} & $0.049$\std{0.028} & $0.96$\std{0.01}\\
\textsc{NeSyDM} & 1 & 40 & $0.997$\std{0.001} & $0.002$\std{0.001} & $0.021$\std{0.009} & $0.93$\std{0.00}\\
 \textsc{Bears} (best fair) & 5 & 20 & $0.997$\std{0.000} & $0.007$\std{0.003} & $0.102$\std{0.063} & $0.91$\std{0.00}\\
\textsc{Bears} (aggressive) & 5 & 20 & $0.678$\std{0.140} & $0.656$\std{0.005} & $0.185$\std{0.093} & $1.65$\std{0.06}\\
\bottomrule
\end{tabular}
\caption{Mixed-identifiability MNIST task: mean$\pm$std over $20$
  seeds (\textsc{Bears} rows: over four replicate five-member ensembles).}
\label{tab:exp}
\end{table}

\paragraph{Results.}
Table~\ref{tab:exp} orders the methods by how they use the symmetry.
The base model, using none, is confidently wrong on the confusable
digits. Orbit-averaging the same twenty models imposes the invariant
point, removing that error while leaving label accuracy and the
identifiable digit unchanged. The RS-independence loss, trained toward
the same target, lands four times higher: gradient descent
approximates what the closed form imposes. \textsc{NeSyDM} comes
within a factor of two only after ten times the training budget and a
configuration search; at the shared budget it is four times worse.
\textsc{Bears} stays roughly eight times above, and its aggressive
setting loses label accuracy without a reliable calibration gain,
because its diversity pressure is global. XOR-parity repeats the
pattern: the base model commits to a convention on over half the
seeds, orbit-averaging is calibrated by construction, and
\textsc{NeSyDM} fails outright on two of twenty seeds
(Appendix~\ref{app:exp}).

\section{Discussion}
\label{sec:discussion}

\subsection{A generating-function consequence}
\label{sec:examples}

The categorified reading has an analytic consequence already in the
simplest relational model. We generalize
Example~\ref{ex:running} to $n$ individuals with one unary predicate, so
a $\sigma$-structure is a subset of $\{S(1),\dots,S(n)\}$,
$\Aut(T)=\Sym(n)$, and the exchangeable belief
$b(\omega)=\beta^{\lvert\omega\rvert}$ is $\Sym(n)$-invariant, so the
symmetry parameter is maximal, $G=\Sym(n)$. Orbits are indexed by the
number $m$ of smokers, with size $\binom{n}{m}$ and stabilizer of
order $m!(n-m)!$, giving
\begin{equation}
\label{eq:exch}
\textstyle\sum_{\omega} b(\omega)=(1+\beta)^{n},\qquad
\lvert\Omega\hquot\Sym(n)\rvert_{b}=\frac{(1+\beta)^{n}}{n!},
\end{equation}
in agreement with Theorem~\ref{thm:main} ($\lvert G\rvert=n!$). The
closed form is machine-checked for every $n$
(Appendix~\ref{app:lean}).

\begin{corollary}[Inference generating functions]
\label{cor:egf}
For the exchangeable model, the flat functional \eqref{eq:wmc}
assembles into the \emph{ordinary} generating function
$\sum_n(1+\beta)^n x^n=\frac{1}{1-(1+\beta)x}$. The belief-weighted
homotopy cardinality \eqref{eq:functional} assembles instead into the
\emph{exponential} generating function
$\sum_n\frac{(1+\beta)^n}{n!}x^n=e^{(1+\beta)x}$.
\end{corollary}

Appendix~\ref{app:related} relates this corollary to the theory of
combinatorial species \citep{joyal1981species}, whose composition
rules explain why the homotopy-cardinality reading composes correctly
for relational models built from parts.

\subsection{Comparison of types and sets}
\label{sec:impact}

The type-theoretic reading is what exposes the symmetry parameter and
the truncation level, and what identifies confusion sets with orbits
(Section~\ref{sec:rs}). The
resulting theorems (Theorem~\ref{thm:main},
Propositions~\ref{prop:inert} and~\ref{prop:method}, and
Corollary~\ref{cor:egf}) deliberately sit in the finite,
$1$-truncated fragment, where the type-based and set-based accounts
agree and the proofs reduce to orbit--stabilizer arithmetic; this is
why the Lean development checks in a proof-irrelevant kernel
(Appendix~\ref{app:lean}). The reading also supplies what the
set-level account cannot name: one object $\sum_{\omega}P(\omega)$
whose decategorifications are the semiring readings of algebraic
model counting.

Two further payoffs are worth naming, and neither is yet realized in a
theorem or an experiment of this paper. The first is proof-relevance:
a set-based truth value cannot distinguish ``provable'' from
``provable in three symmetric ways'', whereas the truncation level of
$P$ puts satisfaction checking, derivation counting, and
symmetric-derivation counting on one axis; our results all sit at the
$1$-truncated level. The second is representation invariance: the
univalence axiom \citep{hottbook} lets equivalent types be treated as
equal (Appendix~\ref{app:hott}), so inference computed in one symbolic
encoding equals inference in any equivalent one with no separate
proof, where set-based foundations re-establish such invariance by
hand. What we machine-check (\texttt{groupoidCard\_transport},
Appendix~\ref{app:lean}) is the finite, set-level instance, not the
univalent statement; Appendix~\ref{app:proofs} spells out the
consequence for description-logic NeSy systems.

\subsection{The continuous side, and an open problem}
\label{sec:cohesion}

Plain homotopy type theory supplies identifications but neither a
metric nor a gradient.  \emph{Real-cohesive} homotopy type
theory \citep{shulman2018brouwer}
equips every type with a synthetic notion of continuity
alongside its identifications; its \emph{shape modality} collapses a
continuous type to its discrete content, a synthetic symbol-grounding
map.

One gap remains: homotopy cardinality is defined for $\pi$-finite types (Section~\ref{sec:sigma}), and a
synthetic measure compatible with cohesion that lets
Eqn.~\eqref{eq:functional} range over a continuous $\Omega$ does not
yet exist. This is the central open problem the framework raises.
Appendix~\ref{app:limitations} collects the other assumptions of the
framework, and what fails when each of them does.

\section{Related work}
\label{sec:related}

Algebraic model counting \citep{kimmig2017algebraic}
reads probabilistic, fuzzy, tropical, and provenance inference
as one semiring sum, which we recover as decategorifications of the
dependent sum, with Scallop \citep{huang2021scallop} the set-valued
case and lifted inference
\citep{vandenbroeck2011lifted,niepert2014tractability} the orbit
machinery of Theorem~\ref{thm:main}. The reasoning-shortcut line
\citep{marconato2023notall,vankrieken2024independence,vankrieken2025rsindependence,marconato2024bears,vankrieken2025nesydm}
supplies confusion sets and uniform targets set-theoretically, and
Section~\ref{sec:rs} reads them as orbits and the invariant point of
a group action. Categorical deep learning
\citep{gavranovic2024categorical} works
under strict equality, below the identity-type structure
we use, and homotopy-theoretic models of
neural data \citep{manin2024homotopy} apply algebraic topology to
neural activity rather than type theory to inference.
Appendix~\ref{app:related} takes each line in turn.

\section{Conclusion}
\label{sec:conclusion}

Read over types rather than sets, neurosymbolic inference becomes
a belief-weighted homotopy cardinality that
remembers structure symmetries and proof identity and recovers the
classical functional when symmetries are trivial
(Theorem~\ref{thm:main}, machine-checked).  The symmetry it exposes is
the one behind reasoning shortcuts, and orbit-averaging, its closed-form
single-model method, is better calibrated than ensemble and
training-time baselines.

\acks{This work was supported by funding from King Abdullah University of
Science and Technology (KAUST), through the KAUST Center of Excellence
for Smart Health (KCSH), under award number 5932, and the KAUST Center
of Excellence for Generative AI, under award number 5940.}

\bibliography{references}

\appendix

\section{A fuller introduction to homotopy type theory}
\label{app:hott}

This appendix expands Section~\ref{sec:hott}. It first fixes the
logical setting (signatures, structures, truth, models) and then the
type-theoretic notions used in the main text; it is self-contained and
follows \citet{hottbook}, to which all numbered results refer, with
illustrations in logic-programming and knowledge-graph terms.

\subsection{Signatures, structures, truth, and models}
A \emph{signature} $\sigma$ is a set of relation, function, and constant symbols, each with
an arity. A \emph{$\sigma$-structure} $\omega$ fixes a domain $D$ of individuals and
interprets every symbol of $\sigma$ over $D$: a $k$-ary relation symbol as a subset of
$D^{k}$, a function symbol as a map $D^{k}\to D$, a constant as an element of $D$. The
$\sigma$-structures over a fixed finite domain form the space $\Omega$. A sentence $\varphi$
is \emph{true} in $\omega$, written $\omega\models\varphi$, by the Tarskian satisfaction
relation; a $\sigma$-structure with $\omega\models\psi$ for every axiom $\psi$ of a theory
$T$ is a \emph{model} of $T$, the least Herbrand model of a definite logic program being the
standard example. Two $\sigma$-structures are \emph{isomorphic} when a renaming of
individuals (a bijection of $D$) carries one to the other while preserving every interpreted
symbol; an isomorphism of a structure with itself is an \emph{automorphism}, and the
automorphisms of $T$ are the renamings under which $T$ is invariant. These are the objects
Section~\ref{sec:hott} promotes one level: $\Omega$ becomes a type whose identifications are
exactly these isomorphisms, and the truth value $\omega\models\varphi$ becomes the proof
family $P(\omega)$ of derivations that witness $\varphi$.

\subsection{Types, terms, and the dependent function and pair types}
Type theory manipulates \emph{types} $A,B,\dots$ and \emph{terms} $a:A$. From a type $A$
and a family $B\colon A\to\U$ (with $\U$ a \emph{universe}, a type of types) one forms the
\emph{dependent function type} $\prod_{a:A}B(a)$ (ordinary $A\to B$ when $B$ is constant)
and the \emph{dependent pair type} $\sum_{a:A}B(a)$ of pairs $(a,b)$ with $b:B(a)$
(ordinary $A\times B$ when $B$ is constant). Under propositions-as-types $\prod$ reads
``for all'' and $\sum$ ``there exists'': $\prod_{i}\mathrm{smokes}(i)$ says every
individual smokes, whereas $\sum_{\omega:\Omega}P(\omega)$ is the type of a $\sigma$-structure
$\omega$ paired with a derivation $p:P(\omega)$ of the query, the (structure, proof) pairs the
inference functional aggregates. When each $B(a)$ is a yes/no fact, $\sum_{a:A}B(a)$ is the
subtype of $a$ satisfying $B$, for instance the models of a theory.

\subsection{Identity types and the groupoid structure}
For $a,b:A$ the \emph{identity type} $a=_A b$ (also $\Id_A(a,b)$) collects the
identifications of $a$ with $b$; for $\sigma$-structures these are the renamings of
individuals carrying one model to the other. Its sole constructor is reflexivity
$\mathrm{refl}_a\colon a=_A a$, and its induction principle (path induction) says that to
define something for all $p\colon a=_A b$ it suffices to treat the case $b\equiv a$,
$p\equiv\mathrm{refl}_a$. Path induction generates the higher structure once $p$ is read
as a \emph{path}: reflexivity is the constant path (the identity renaming), symmetry
$p^{-1}$ the inverse renaming, and transitivity $p\cdot q$ their composition
(\citealp[\S2.1]{hottbook}). These operations satisfy the groupoid laws up to higher
paths, so each type is a (weak) $\infty$-groupoid.
\begin{center}\small
\begin{tabular}{@{}lll@{}}
\toprule
equality & homotopy & $\infty$-groupoid\\
\midrule
reflexivity & constant path & identity morphism\\
symmetry & path inversion & inverse morphism\\
transitivity & path concatenation & composition\\
\bottomrule
\end{tabular}
\end{center}
The self-identifications $a=_A a$ form, under composition, the automorphism group
$\Aut(a)$ of the model; the nontrivial automorphisms of a knowledge graph (relabelings of
its entities that preserve every triple) are exactly its nontrivial elements. A set is the
degenerate case in which every such group is trivial.

\subsection{Equivalence, univalence, and transport}
A function $f\colon A\to B$ is an \emph{equivalence}, $A\simeq B$, when it has a coherent
inverse (\citealp[\S2.4,\S4]{hottbook}). For $A,B:\U$ there is a canonical
$\mathsf{idtoeqv}\colon (A=_\U B)\to(A\simeq B)$, and the \emph{univalence axiom}
(\citealp[Axiom 2.10.3]{hottbook}) makes it an equivalence, so $(A=_\U B)\simeq(A\simeq B)$:
equivalent types may be identified. Its computational content is \emph{transport}, and
along $\mathsf{ua}(f)$ transport computes to $f$ (\citealp[\S2.10]{hottbook}). Concretely,
two knowledge bases differing only by an isomorphic renaming of their vocabulary (distinct
IRIs or constant names for the same entities) have identified structure types, and the
inference functional transports between them unchanged; representation invariance
(Section~\ref{sec:impact}) is automatic rather than proved per encoding.

\subsection{Truncation levels (h-levels)}
The amount of identification structure a type carries is its \emph{truncation level},
defined by recursion (\citealp[Def.~7.1.1]{hottbook}): $X$ is a $(-2)$-type if it is
contractible, and an $(n{+}1)$-type if $x=_X y$ is an $n$-type for all $x,y:X$. Therefore
$(-1)$-types are the \emph{mere propositions} (any two elements identified;
\citealp[Def.~3.3.1]{hottbook}), the truth of a ground atom; $0$-types are the \emph{sets}
(any two parallel paths identified; \citealp[Def.~3.1.1]{hottbook}), the
$\sigma$-structures; and $1$-types are the \emph{groupoids}, $\sigma$-structures with their
renaming symmetries. The levels are cumulative (\citealp[Thm.~7.1.7]{hottbook}) and closed
under $\sum$ and $\prod$ (\citealp[Thm.~7.1.8--7.1.9]{hottbook}), so the structure--witness
object $\sum_\omega P(\omega)$ has a level determined by those of $\Omega$ and $P$.
Truncating to level $n$ forgets all structure above $n$.

\subsection{Homotopy cardinality}
For a $\pi$-finite groupoid (finitely many isomorphism classes, finite automorphism
groups) the homotopy (or groupoid) cardinality is given by Eqn.~\eqref{eq:hcard}
\citep{baez2001finite,leinster2008euler}. The two facts we use are that a set of $k$
$\sigma$-structures has cardinality $k$, and that for a finite renaming group $G$ acting on a
finite set $X$ of $\sigma$-structures the action groupoid satisfies
$\lvert X\hquot G\rvert=\frac{\lvert X\rvert}{\lvert G\rvert}$ by orbit--stabilizer: the count of
models up to isomorphism. Homotopy cardinality is the decategorification under which
disjoint unions add, products multiply, and the groupoid of finite structures of a species
has the species' exponential generating function as its cardinality, the fact behind
Corollary~\ref{cor:egf}.

\begin{remark}[Scope of the valuation]
\label{rem:divisible}
The weight $\tfrac{1}{\lvert\Aut\rvert}$ asks the semiring to divide by
group orders, so Eqn.~\eqref{eq:functional} is semiring-valued only where
that makes sense. Two cases qualify: $(+,\times)$ over $\mathbb{Q}^{+}$ or
$\mathbb{R}^{+}$ (probability, gradients), where $\tfrac{1}{\lvert G\rvert}$
exists literally, and the idempotent semirings, where $\lvert G\rvert\cdot 1=1$
renders it the identity (Proposition~\ref{prop:inert}(b)). It
\emph{leaves the carrier} for integer counting: the symmetry-corrected
$\#$SAT value $\tfrac{1}{\lvert G\rvert}\sum_{x} b(x)$ is in general
rational, not in $\mathbb{N}$, so $\#$SAT is non-idempotent yet still
cannot see the correction as an integer count. The number of
$\sigma$-structures up to isomorphism is a rational invariant even when
the flat count is an integer, the same rationality behind the
exponential generating function of Section~\ref{sec:examples}: homotopy
cardinality is intrinsically a $\mathbb{Q}$-linear notion of size.
\end{remark}

\section{Proofs}
\label{app:proofs}

\paragraph{Theorem~\ref{thm:main}.}
By orbit--stabilizer each orbit $[x]$ has $\frac{\lvert G\rvert}{\lvert\Stab_G(x)\rvert}$
members, all of equal belief by $G$-invariance, so
\[
\sum_{x\in X}b(x)=\sum_{[x]}\frac{\lvert G\rvert}{\lvert\Stab_G(x)\rvert}\,b(x)
=\lvert G\rvert\sum_{[x]}\frac{b(x)}{\lvert\Stab_G(x)\rvert}
=\lvert G\rvert\cdot\lvert X\hquot G\rvert_{b}.
\]
Dividing by $\lvert G\rvert$ gives the claim. The argument is formalized in Lean~4
(Appendix~\ref{app:lean}).

\paragraph{Proposition~\ref{prop:inert}(b).}
Read the dependent sum in a commutative semiring with idempotent
$\oplus$. On each orbit $[x]$ the belief is constant, $b(\cdot)=v$, by
$G$-invariance. The set-based valuation of the orbit is
$\bigoplus_{y\in[x]} v=v$ (idempotence: any finite $\oplus$ of copies of
$v$ is $v$), independent of the orbit size; and idempotence makes
$\lvert G\rvert\cdot 1=1$, so the weight $\tfrac{1}{\lvert\Stab\rvert}$
is the identity and the type-based valuation of the orbit is $v$ as
well. The two agree orbit by orbit, so
$\lvert X\hquot G\rvert_{b}=\bigoplus_{x\in X}b(x)$ for every $G$.
Case (a) is Theorem~\ref{thm:main} at $G=\mathbf 1$.

\paragraph{Proposition~\ref{prop:method}(i), accuracy preservation.}
Fix a label $y$ and write $F=\beta^{-1}(y)$ for its fiber. Each $g\in G$ preserves
$\beta$, so $g^{-1}$ restricts to a bijection of $F$. Hence
\[
\begin{aligned}
\sum_{c\in F}\bar p(c\mid x)
&=\frac{1}{\lvert G\rvert}\sum_{g\in G}\sum_{c\in F}p_\theta(g^{-1}\!\cdot c\mid x)\\
&=\frac{1}{\lvert G\rvert}\sum_{g\in G}\sum_{c'\in F}p_\theta(c'\mid x)
=\sum_{c\in F}p_\theta(c\mid x),
\end{aligned}
\]
the inner sum being reindexed by $c'=g^{-1}\!\cdot c$. The induced label distribution is
unchanged, so any label decision (and its accuracy) is preserved. Part~(ii) is the
orbit-conditional uniformity machine-checked as \texttt{invariant\_posterior\_uniform}.

\paragraph{Corollary~\ref{cor:egf}.}
Summing Eqn.~\eqref{eq:exch} against $x^n$ gives $\sum_n(1+\beta)^n x^n=\frac{1}{1-(1+\beta)x}$ for
the flat count and $\sum_n \frac{(1+\beta)^n}{n!}x^n=e^{(1+\beta)x}$ for the homotopy
cardinality, the latter by the exponential series.

\paragraph{Description-logic symmetries.}
In sROIQ a symmetric role declaration $\Sym(R)$ identifies $R(a,b)$ with $R(b,a)$, and
interchangeable nominals make a swap $(a\,b)$ an automorphism of the theory $T$, so
$\lvert\Aut(T)\rvert>1$. A knowledge-compilation circuit for WMC over a grounding of $T$
then carries, unchanged, both $\frac{\mathrm{WMC}}{\lvert\Aut(T)\rvert}$ (Theorem~\ref{thm:main})
and the tropical $(\max,+)$ reading.

\section{The Lean formalization}
\label{app:lean}

The development (\texttt{HottNeSy/GroupoidCard.lean}, Mathlib v4.29.1) compiles with
\texttt{\#print axioms} reporting only \texttt{propext}, \texttt{Classical.choice}, and
\texttt{Quot.sound}, and with no \texttt{sorry}. The lemma \texttt{groupoidCard\_eq} is
Theorem~\ref{thm:main} and \texttt{groupoidCard\_\allowbreak trivial} the
trivial-symmetry recovery. Representation invariance is
\texttt{groupoidCard\_\allowbreak transport}, and
\texttt{invariant\_\allowbreak posterior\_\allowbreak uniform} states that the
orbit-conditional posterior is uniform, the awareness target of
Section~\ref{sec:rs}. The closed form $\frac{(1+\beta)^n}{n!}$ of \eqref{eq:exch} is
verified for every $n$ by \texttt{exch\_\allowbreak groupoidCard}, with supporting
\texttt{exch\_invariant} and \texttt{exch\_flat\_sum}. Finally,
\texttt{groupoidCard\_\allowbreak idempotent\_\allowbreak inert} is
Proposition~\ref{prop:inert}(b): for a $G$-invariant belief into an
idempotent additive monoid the type-based and set-based valuations
coincide for every $G$, with supporting \texttt{nsmul\_\allowbreak idem} and
\texttt{sum\_\allowbreak eq\_\allowbreak sum\_\allowbreak orbits\_\allowbreak nsmul}. The whole argument is
orbit--stabilizer arithmetic, so the finite case needs no homotopy type theory and checks
in a proof-irrelevant kernel.

\section{Experimental details}
\label{app:exp}

\paragraph{Architectures and training.}
Perception is a small convolutional network (two conv layers, two linear layers) mapping
a $28\times28$ image to digit logits, trained with Adam ($10^{-3}$, batch $256$) on the
merged label by minimizing the negative log-likelihood of $\beta$, the standard NeSy
objective. The symmetry group $G$ is enumerated from $\beta$ as the product of symmetric
groups on its label fibers. Orbit-averaging \eqref{eq:reynolds} is applied at inference.
\textsc{Bears} trains $K=5$ members sequentially; member $t$ adds $-\gamma_1\,
\mathrm{KL}(p_t\Vert\bar p_{<t})-\gamma_2 H(p_t)$ to the loss (diversity from the running
average plus entropy), a variant of Eq.~7 of \citet{marconato2024bears} in which the
running average is over the previous members and their normalization constants are
omitted, and the ensemble averages members.

\paragraph{Metrics.}
All calibration numbers are expected calibration error (ECE): test examples are binned by
their top-1 confidence, and ECE is the weighted mean over bins of the gap
$\lvert\text{accuracy}-\text{confidence}\rvert$. On the mixed task, accuracy is
\emph{tie-aware}: a top-1 tie
among $k$ classes counts as $1/k$ correct (expected top-1 correctness under uniform
tie-breaking), so a deliberately uniform posterior is scored by its expected
correctness rather than by an arbitrary argmax. We report ECE on the identifiable digit (\emph{id-ECE}) and on the
confusable digits (\emph{conf-ECE}) separately, the accuracy of the merged label $\beta$
(\emph{label}), and the mean entropy of the digit posterior on the confusable digits
(\emph{conf-ent.}), which is higher when the model is more uncertain, as the aware target
requires. On the XOR task \emph{bit-ECE} is the same ECE applied to the posterior over a
parity bit, with plain argmax accuracy; the missing tie adjustment can only penalize
orbit-averaging, whose parity marginal ties at exactly $\tfrac12$.

\paragraph{Mixed task.}
Table~\ref{tab:exp} reports means over $20$ base seeds for the base and orbit-averaged
rows; each \textsc{Bears} row is the mean over four replicate five-member ensembles
built from disjoint blocks of member seeds ($100$--$119$). The two rows are the ends of
a $(\gamma_1,\gamma_2)$ sweep: the best label-preserving setting
($\gamma_1{=}0.1$, $\gamma_2{=}0$) and the aggressive one
($\gamma_1{=}0.5$, $\gamma_2{=}0.1$). Replication is what exposes the aggressive
setting: its conf-ECE runs from $0.097$ to $0.314$ across the four replicates, so
pushing diversity harder does not trade accuracy for calibration; it varies.

\paragraph{RS-independence on the mixed task.}
The regularizer of \citet{vankrieken2025rsindependence}, their Eq.~7, replaces the label
likelihood by the KL divergence from the uniform distribution on the confusion set of the
observed label to the concept posterior, so it trains one model toward the same target
orbit-averaging imposes. We reimplemented it from their released code, with the
uniform targets derived from the label fibers of $\beta$. Table~\ref{tab:exp} reports
it at the budget of the other single-model rows; longer training and the
hyperparameters of the \citet{vankrieken2025rsindependence} paper stay within one
standard deviation of that row. On this task the label fiber of $\beta$ is exactly an orbit of $G$,
so their loss trains toward the invariant point Eqn.~\eqref{eq:reynolds} computes in
closed form, but it needs the confusion set during training and a retraining run,
whereas orbit-averaging needs the group only at inference and leaves the trained
model untouched.

\paragraph{NeSyDM.}
Neurosymbolic diffusion models \citep{vankrieken2025nesydm} replace the factorized
perception by a discrete diffusion process that unmasks the concepts a subset at a time,
so one model can represent the uniform posterior on a confusion set without an ensemble.
We reimplemented the simple variant their paper reports from the official
implementation. One deviation is deliberate: upstream estimates the output loss by
REINFORCE leave-one-out sampling because weighted model counting is \#P-hard in
general, and we enumerate the expectation exactly, which never hurts the baseline at
matched configurations. Table~\ref{tab:exp} reports the best label-preserving
configuration of a $21$-configuration sweep per task, trained for $40$ epochs on the
mixed task and $100$ on XOR against the $4$ and $5$ of every other method, with test
posteriors read off $1000$ sampler draws per input; the budget-matched row changes
only the epoch count, with no re-selection. At the converged budget its confidence
matches its tie-aware accuracy on the confusable digits ($0.421$ against $0.407$),
the signature of a single model at the invariant point.

\paragraph{XOR-parity task.}
Each input is a pair of MNIST images. The two concepts $b_1$ and $b_2$ are the
parities of the digits shown, the label is their exclusive or,
$y=b_1\oplus b_2$, and training supervises $y$ alone. The global flip $(b_1,b_2)\mapsto(1{-}b_1,1{-}b_2)$ preserves XOR, so
$G=\mathbb{Z}/2$ and each parity is identifiable only up to the flip. Over $20$ seeds the
base model commits to the wrong convention on $55\%$ of them (bit-ECE $0.539$, confidence
$0.98$); orbit-averaging a single model gives bit-ECE $0.013\pm0.024$ with the label and the
relation $b_1\!=\!b_2$ accuracy unchanged. A naive ensemble inherits the majority
convention's bias: over the two disjoint ten-member ensembles that $20$ seeds allow, its
bit-ECE is $0.589$ and $0.005$, the second only because that block of seeds happens to
split evenly between the two conventions. An ensemble balanced across conventions with the
help of concept ground truth, which the other methods do not receive, only approximates
the invariant point (bit-ECE $0.165$ and $0.005$ on the same two blocks) and needs many
models. Doubling to a single twenty-member ensemble does not close the gap (bit-ECE
$0.532$ naive, $0.084$ balanced).

NeSyDM stays closer on average to the invariant point than either ensemble and further
from it than orbit-averaging. Over the same $20$ seeds its selected configuration reaches bit-ECE
$0.060\pm0.055$ at label accuracy $0.936\pm0.146$ and bit entropy $0.692$, against the
maximum $0.693$, so entropy alone again fails to separate the methods; calibration
and label accuracy do. The label spread is not seed noise around a solved
task: on $2$ of the $20$ seeds the model never learns XOR at all, even at the full
$100$-epoch budget (label accuracy $0.50$), whereas Eqn.~\eqref{eq:reynolds} imposes the
invariant point rather than approaching it by training. At the matched budget of $5$
epochs, $12$ of $20$ seeds never learn the task (label accuracy $0.504$) and are
calibrated only by being uniform, while the eight that do learn reach bit-ECE $0.140$:
here too label accuracy, not calibration, separates awareness from
a model that has not learned the task.

The same task is a control for the RS-independence loss. This perception factorizes over
the two parities, $p(b_1,b_2\mid x)=f(b_1\mid x_1)f(b_2\mid x_2)$, which is the model
class Theorem~7 of \citet{vankrieken2025rsindependence} rules out for XOR, and their
Eq.~7 collapses on it exactly as that theorem predicts. Every parity marginal converges to
the constant $\tfrac12$ (mean $0.500$ with a spread of $6\times10^{-4}$ across
the test set), and label accuracy is then $0.501\pm0.008$ over the same $20$ seeds, chance
level, against $0.970$ for orbit-averaging on the same architecture. The
hyperparameters of their paper give the same collapse. What matters for
reading the numbers is that the collapsed model's bit-ECE ($0.016\pm0.012$) and bit
entropy ($0.693$, the maximum) are indistinguishable from orbit-averaging ($0.013$
and $0.693$): only label accuracy separates a shortcut-aware model from one
that learned nothing.

\section{Misspecified symmetry groups}
\label{app:ablation}

Section~\ref{sec:exp} hands orbit-averaging the true group, so we measured what a wrong
group costs. We reused the twenty base models of Table~\ref{tab:exp} without retraining and
symmetrized each of them with nine different groups, so the rows of
Table~\ref{tab:ablation} are paired on the same perceptions and differ only in $G$. The
groups fall into four kinds: the trivial group (no averaging), the true group, proper
subgroups of the true group (\emph{under-specification}), supergroups that merge concepts
the knowledge separates (\emph{over-specification}), and groups that permute concepts
across the label fibers of $\beta$ (\emph{cross-fiber}). Table~\ref{tab:ablation} reports
seven of the nine variants; the two omitted ones, $\Sym\{2,3\}$ and $\Sym\{2,3,4\}$,
are cross-fiber groups that also cost label accuracy.

\begin{table}[t]
\centering\small
\begin{tabular}{@{}llcccc@{}}
\toprule
group $G$ & kind & $\lvert G\rvert$ & label & id-ECE & conf-ECE\\
\midrule
$\mathbf{1}$ & base & $1$ & $0.996$\std{0.001} & $0.002$\std{0.001} & $0.588$\std{0.109}\\
$\Sym\{1,2\}\times\Sym\{3,4,5\}$ & true & $12$ & $0.996$\std{0.001} & $0.004$\std{0.002} & $0.013$\std{0.008}\\
$\Sym\{1,2\}$ & under & $2$ & $0.996$\std{0.001} & $0.003$\std{0.001} & $0.379$\std{0.059}\\
$\Sym\{3,4,5\}$ & under & $6$ & $0.996$\std{0.001} & $0.004$\std{0.002} & $0.216$\std{0.108}\\
$\Sym\{0,1,2\}\times\Sym\{3,4,5\}$ & over & $36$ & $0.835$\std{0.001} & $0.055$\std{0.037} & $0.009$\std{0.006}\\
$\Sym\{0,1,2,3,4,5\}$ & over & $720$ & $0.478$\std{0.000} & $0.101$\std{0.062} & $0.022$\std{0.009}\\
$\Sym\{0,3\}$ & cross-fiber & $2$ & $0.842$\std{0.016} & $0.003$\std{0.002} & $0.537$\std{0.136}\\
\bottomrule
\end{tabular}
\caption{Orbit-averaging under a misspecified group on the mixed-identifiability task
  (mean$\pm$std over the same $20$ base models). The base and
  true-group rows reproduce the corresponding rows of Table~\ref{tab:exp}, which are
  computed from the same twenty perceptions. Label accuracy separates the correct group from an incorrect one; conf-ECE does
  not.}
\label{tab:ablation}
\end{table}

\paragraph{Under-specification is safe and partial.}
A proper subgroup of the true $G$ is still output-preserving, so
Proposition~\ref{prop:method}(i) applies and label accuracy stays at the base value in
both under-specified rows. Such a subgroup calibrates the orbits it covers and
leaves the rest untouched: $\Sym\{1,2\}$ alone takes the ECE on the
confusable pair $\{1,2\}$ from $0.492$ to $0.005$ while the ECE on the confusable triple
$\{3,4,5\}$ stays at its base value, and $\Sym\{3,4,5\}$ alone mirrors this. The
aggregate conf-ECE of an under-specified row is therefore a mixture of one corrected
block and one untouched block rather than a partial correction of both.

\paragraph{Over-specification and cross-fiber groups cost label accuracy.}
A group larger than the true one is no longer output-preserving. Merging the
identifiable digit $0$ into a confusion class costs label accuracy and drops accuracy
on that digit from $0.997$ to $0.373$; the full $\Sym\{0,\dots,5\}$ halves label
accuracy outright. The cross-fiber group $\Sym\{0,3\}$ likewise costs label accuracy
and halves the identifiable-digit accuracy while leaving calibration where it was, so
it pays the price of a wrong group and buys nothing.

\paragraph{Label accuracy is the detector, not calibration error.}
Both over-specified rows keep a low conf-ECE, one of them below the
true group's, because a uniform posterior over a larger orbit stays self-consistent
under the tie-aware scoring of Appendix~\ref{app:exp}: a confidence of $1/k$ is matched by
an expected correctness of $1/k$ whenever the true concept lies in the orbit. A low
calibration error alone therefore does not certify that $G$ is correct. Validation label
accuracy does, because Proposition~\ref{prop:method}(i) preserves it exactly whenever $G$
is output-preserving, so any drop is evidence against the assumed group. This makes
misspecification a checkable modelling error rather than a silent one: $G$ is read off the
knowledge $\beta$ (Section~\ref{sec:rs}) and not estimated from data, and the check needs
only the validation labels the model is already trained on.

\section{Limitations}
\label{app:limitations}

\paragraph{Exact and partial symmetry.}
The closed form of Proposition~\ref{prop:method} is exact when $G$ is a genuine symmetry
of the task, so that every relabeling in $G$ leaves the program output unchanged on every
input. Many problems carry only an approximate symmetry: a relabeling may preserve the
output on most inputs but not on all of them, or the data may carry weak evidence that
distinguishes two concepts a purely logical reading would identify. Orbit-averaging then
returns the uniform distribution on the orbit while the correct posterior is near uniform
but not uniform, and that residual is what the method cannot represent.
Section~\ref{sec:rs} notes that an ensemble \citep{marconato2024bears} or an expressive
joint model \citep{vankrieken2025nesydm} covers the rest of the simplex. A hybrid that
keeps the closed-form invariant component and learns the residual is future work.

\paragraph{Belief invariance.}
Theorem~\ref{thm:main} assumes a $G$-invariant belief, so the symmetry parameter is the
symmetry shared by the theory and the perception rather than the theory's symmetry alone
(Section~\ref{sec:lift}). A perception that scores two individuals differently breaks
their symmetry and forces $G=\mathbf{1}$. The lift is then inert by
Proposition~\ref{prop:inert}(a) and orbit-averaging is the identity map. This is the
behavior we want, because evidence that separates members of a confusion set should not
be averaged away, but it does bound the scope of the method to the symmetry that survives
perception. Which part of a symmetry the data actually breaks is a modelling decision we
do not automate.

\paragraph{Known and estimated $G$.}
Both tasks of Section~\ref{sec:exp} hand the method the true group, because the knowledge
$\beta$ is given and $G$ is enumerated from it. Section~\ref{sec:rs} notes that this
enumeration is graph-automorphism detection on the theory, for which practical solvers
exist \citep{mckay2014practical,darga2008faster}. Those solvers return a symmetry of the
\emph{knowledge}. Estimation from \emph{data} is a different matter, needed when the
knowledge is unavailable or when the data breaks a symmetry the knowledge has, and we do
not address it. Misspecification costs something in both directions: a group smaller
than the true one leaves part of each confusion set overconfident, and a group larger
than it breaks the hypothesis of Proposition~\ref{prop:method}(i), so the method no
longer preserves label accuracy. Appendix~\ref{app:ablation} measures both directions
on the mixed task and shows that validation label accuracy detects the second case.

\paragraph{Scale of the evidence.}
Our experiments are at MNIST scale, with a handful of concepts and a group small enough
to enumerate, and two consequences follow. First, Eqn.~\eqref{eq:reynolds} is written as
a literal sum over $G$; a group whose order grows factorially with the domain, as
$\Sym(n)$ does in Section~\ref{sec:examples}, needs the orbit structure instead of the
sum, and we have not implemented that case. Second, the evidence says nothing yet about
relational tasks, knowledge graphs, or concept-bottleneck models at scale, where
confusion sets are larger and the group is harder to obtain. We defer those benchmarks to
future work.

\paragraph{Idempotent semirings.}
Proposition~\ref{prop:inert}(b) is a limitation as much as a result. On the idempotent
semirings, which include fuzzy $(\max,\min)$ and MPE or Viterbi $(\max,+)$, the lifted
and set-based values agree for every $G$, so the framework says nothing new about systems
built on them. The refinement is confined to the non-idempotent semirings.

\paragraph{The continuous case.}
Eqn.~\eqref{eq:deraedt} covers weighted model integration through the base measure $m$,
and the lifted functional does not. Homotopy cardinality is defined for $\pi$-finite
types (Section~\ref{sec:sigma}), and a synthetic measure compatible with cohesion that
would let Eqn.~\eqref{eq:functional} range over a continuous $\Omega$ does not yet exist
(Section~\ref{sec:cohesion}). Every result in this paper is therefore finite, and the
continuous half of neurosymbolic inference stays outside the framework.

\section{Related work in more detail}
\label{app:related}

This appendix expands Section~\ref{sec:related}, adding no references beyond those
already cited there.

\paragraph{Algebraic model counting.}
Algebraic model counting \citep{kimmig2017algebraic} reads probabilistic, fuzzy,
tropical, and provenance inference as one sum over models in a commutative semiring,
parameterized by $(\oplus,\otimes)$. Our value object keeps the models and their
derivations as a type and applies the semiring afterwards, as the valuation
$\lVert\cdot\rVert_V$ of Eqn.~\eqref{eq:functional}. The difference is what the
parameterization ranges over. For algebraic model counting the semiring is the only
parameter, whereas here it is one of three, alongside the symmetry parameter and the
truncation level. Automorphisms have no place in the set-based sum, which is why the
correction of Theorem~\ref{thm:main} cannot be obtained by choosing a different semiring.
Proposition~\ref{prop:inert}(b) bounds the difference exactly: on the idempotent
semirings the two readings agree. Scallop \citep{huang2021scallop} is the set-valued
case, where the proof object records which derivations witness a query rather than only
whether one exists, which is our $P$ at truncation level $0$; it does not identify
derivations related by a symmetry.

\paragraph{Lifted inference.}
Lifted inference \citep{vandenbroeck2011lifted,niepert2014tractability} uses the same
orbits we do, for a different purpose. It exploits symmetry to evaluate the flat sum
$\sum_{x}b(x)$ faster, and the number it returns is the set-based one. Our symmetry
parameter changes which number counts as the answer, and Theorem~\ref{thm:main} states
that the two differ by the factor $\lvert G\rvert$ whenever the belief is $G$-invariant.
The exchangeability that makes lifted inference tractable
\citep{niepert2014tractability} is the invariance Corollary~\ref{cor:egf} uses: the
ordinary generating function is the flat count and the exponential one is the quotient,
so the two views differ by a decategorification rather than by an approximation.

\paragraph{The reasoning-shortcut line.}
This line begins with \citet{marconato2023notall}, who show that the loss-minimizing
perceptions are convex combinations of the deterministic optima reached by
output-preserving relabelings. Then \citet{vankrieken2024independence} identify the
independence assumption as what drives a model to commit to one of those optima, and
\citet{vankrieken2025rsindependence} name the confusion set and regularize toward a
uniform target on it. The ensemble of \citet{marconato2024bears} reaches the same target
by training members to disagree, and \citet{vankrieken2025nesydm} reach it by fitting an
expressive joint distribution over concepts. All three approach the target from outside,
by adding models or capacity until the required distribution can be represented. The
orbit reading gives it directly: a confusion set is an orbit, the uniform distribution is
the only invariant point of the group action on its simplex
(Proposition~\ref{prop:method}), and Eqn.~\eqref{eq:reynolds} reaches that point in one
forward pass. The assumption moves rather than disappears. \textsc{Bears} and NeSyDM need
no knowledge of $G$ and pay for that in models or in capacity, while orbit-averaging
needs $G$ and pays nothing at training time. Work on identifiability in concept-based
models \citep{bortolotti2025identifiability} studies when concepts are recoverable at
all, and the orbit size $\lvert G\rvert/\lvert\Stab\rvert$ is a per-input measure of the
same degeneracy. The benchmark suite \citep{bortolotti2024rsbench} supplies the task
style our experiments follow.

\paragraph{Categorical deep learning.}
Categorical deep learning \citep{gavranovic2024categorical} gives an algebraic theory of
neural architectures, working $1$- and $2$-categorically with strict equality between
objects. The equivariance it imposes constrains the maps between representations, whereas
our symmetry parameter constrains the value the inference functional assigns. Strict
equality cannot state internally that two structures are the same up to a chosen
symmetry, which is what the identity type does, and it carries no truncation axis.
Neither line subsumes the other: theirs organizes architectures, ours organizes the
inference functional. Homotopy-theoretic models of neural data \citep{manin2024homotopy}
apply algebraic topology to neural activity, so their objects are spaces built from
recordings rather than types of $\sigma$-structures; the overlap with our work is in the
vocabulary rather than in the constructions.

\paragraph{Combinatorial species.}
In the theory of combinatorial species \citep{joyal1981species}, a
species' exponential generating function is the homotopy cardinality
of its associated groupoid \citep{baez2001finite}, and the
exchangeable model of Section~\ref{sec:examples} is the species
``subsets of the domain''. Because species compose, the
homotopy-cardinality reading composes correctly when relational models
are built from parts, while the flat count overcounts relabelings; the
correct generating function follows from the inference functional
itself, without a separate exchangeability argument.

\end{document}